\def\year{2022}\relax
\documentclass[letterpaper]{article} 
\usepackage{aaai22}  
\usepackage{times}  
\usepackage{helvet}  
\usepackage{courier}  
\usepackage[hyphens]{url}  
\usepackage{graphicx} 
\usepackage{subfigure}
\usepackage{natbib}  
\usepackage{caption} 
\DeclareCaptionStyle{ruled}{labelfont=normalfont,labelsep=colon,strut=off} 
\usepackage{algorithm}
\usepackage{algcompatible}
\usepackage{amsmath}
\usepackage{amssymb}
\usepackage{multirow}
\usepackage{newfloat}
\usepackage{listings}
\floatstyle{ruled}
\newfloat{listing}{tb}{lst}{}
\floatname{listing}{Listing}
\title{CoDiM: Learning with Noisy Labels via Contrastive Semi-Supervised Learning}

\author{
    Xin Zhang\textsuperscript{\rm 1}\equalcontrib,
    Zixuan Liu\textsuperscript{\rm 2}\equalcontrib, 
    Kaiwen Xiao\textsuperscript{\rm 3}, 
    Tian Shen\textsuperscript{\rm 3}, 
    Junzhou Huang\textsuperscript{\rm 3}, \\
    Wei Yang\textsuperscript{\rm 3}, 
    Dimitris Samaras\textsuperscript{\rm 1}, 
    Xiao Han\textsuperscript{\rm 3}\thanks{Corresponding Author}
    }
\affiliations{
    \textsuperscript{\rm 1} Stony Brook University, USA\\
    \textsuperscript{\rm 2} University of Washington, USA\\
    \textsuperscript{\rm 3} Tencent AI Lab, China\\
    xin.zhang@cs.stonybrook.edu, zucksliu@cs.washington.edu

}

\begin{document}

\maketitle

\begin{abstract}
Labels are costly and sometimes unreliable. Noisy label learning, semi-supervised learning, and contrastive learning are three different strategies for designing learning processes requiring less annotation cost. Semi-supervised learning and contrastive learning have been recently demonstrated to improve learning strategies that address datasets with noisy labels. Still, the inner connections between these fields as well as the potential to combine their strengths together have only started to emerge. In this paper, we explore further ways and advantages to fuse them. Specifically, we propose CSSL, a unified \textbf{C}ontrastive \textbf{S}emi-\textbf{S}upervised \textbf{L}earning algorithm, and CoDiM (\textbf{Co}ntrastive \textbf{Di}vide\textbf{M}ix), a novel algorithm for learning with noisy labels. CSSL leverages the power of classical semi-supervised learning and contrastive learning technologies and is further adapted to CoDiM, which learns robustly from multiple types and levels of label noise. We show that CoDiM brings consistent improvements and achieves state-of-the-art results on multiple benchmarks.
\end{abstract}

\section{Introduction}
\begin{figure*}[t]
\centering
\includegraphics[width=0.8\linewidth,height=6cm]{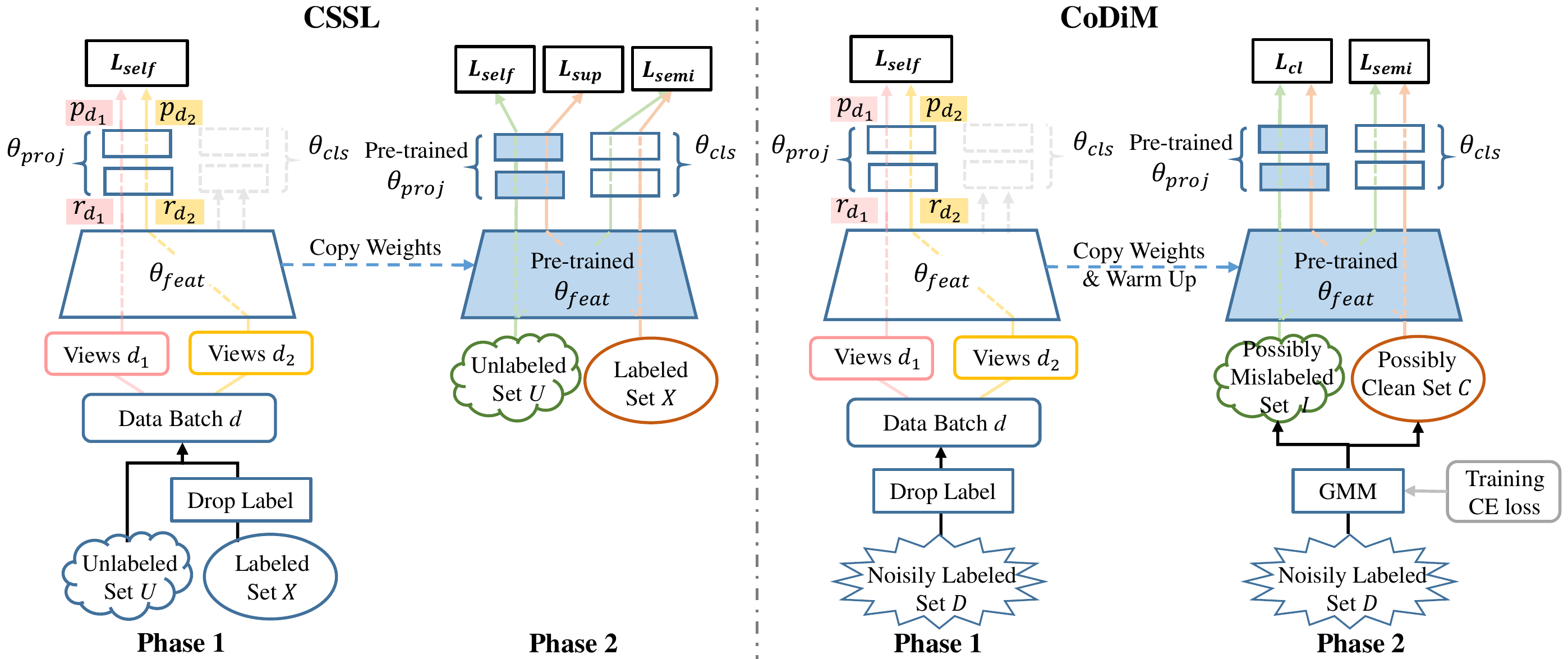}
\caption{The overall framework of CSSL and CoDiM. The former deals with a Semi-Supervised Learning (SSL) task where labels of the labeled set are assumed clean, while the latter is designed to handle a Learning with Noisy Labels (LNL) problem in which a part of training samples are mislabeled. In the first phase, both CSSL and CoDiM regard the whole training set as unlabeled and learn representations by self-supervised contrastive learning (SelfCon). In the second phase, CSSL employs SelfCon loss to the unlabeled set and SupCon loss to the labeled set along with a standard SSL algorithm; CoDiM splits a possibly clean set from the noisy set based on small-loss criterion, discards labels of the noisy set and then applies SSL in a similar manner like CSSL. Note only SupCon or SelfCon loss for the clean set is considered in CoDiM. For brevity, we omit augmentation strategies and regularization rules used in the second phase.}
\label{fig1}
\end{figure*}

Deep learning methods with annotated label supervision have achieved great success in recent years \cite{he2016deep,tan2019efficientnet}, whereas obtaining high-quality label annotations is usually difficult due to constraints on time and labor cost, or the lack of domain knowledge \cite{cheplygina2019not}. Many alternative efforts have been made to detour such expensive processes by developing automated labeling techniques or mining large-scale data with labels through web searching, introducing label noise inevitably, and thus leading models to learn from bias. Furthermore, recent studies have claimed the severity of the over-fitting problem of deep neural networks caused by noisy label bias \cite{zhang2016understanding}, which downgrades the model performance significantly. All of these suggest the necessity and importance to develop methods that could Learn with Noisy Labels (LNL).\\
Enormous researches have been studied to deal with noisy labels. Inspired by recent improvements achieved by Semi-Supervised Learning (SSL) techniques \cite{berthelot2019mixmatch,berthelot2019remixmatch,sohn2020fixmatch}, some methods \cite{arazo2019unsupervised,li2020dividemix} address the potential of designing LNL algorithms in an iterative noise detection \& semi-supervised learning manner. However, performances of these methods will downgrade under scenarios with high ratio label noise. Recently, Contrastive Learning (CL) approaches \cite{chen2020simple,he2020momentum,chen2020big,chen2020improved} have shown great potential on learning good representations by learning a feature extractor and a projector where in projection space, similar samples will be closer while dissimilar samples will be far apart. Seeing its potential on feature learning, some methods try to utilize contrastive learning to help to learn with high ratio noisy labels, by using it to learn a good network initialization \cite{zheltonozhskii2021contrast} or an unsupervised pre-trained label corrector \cite{zhang2020decoupling}. Nevertheless, such methods fail to further utilize contrastive learning techniques. This is mainly due to the lack of exploration on designing and evaluating methods that could better combine CL and SSL together. Furthermore, better ways to strengthen SSL-style LNL methods with contrastive learning techniques need to be explored. \\
In this work, we present CSSL, a simple yet general Contrastive Semi-Supervised Learning algorithm, and CoDiM, a novel learning with noisy labels framework combining the advantages of contrastive learning and SSL-style LNL methods in a more harmonious way. The overall framework of these two algorithms are illustrated in Fig. \ref{fig1}. The key contributions of our work are:
\begin{itemize}
    \item We design a new algorithm named CSSL, which has a self-supervised pre-training phase and a sequential jointly contrastive and semi-supervised learning phase via multi-task learning and address its effectiveness on providing extra consistency regularization. 
    \item We adapt CSSL to CoDiM with several simple yet critical modifications inspired by the state-of-the-art LNL algorithm. We further address the advantage of maintaining a self-supervised/supervised contrastive learning regularization when learning with noisy labels.  
    \item Experimentally, we show that CSSL can bring improvements through learning better representations on semi-supervised learning tasks. We further present extensive experimental results on multiple synthetic and real-world noisy label learning benchmarks and show that CoDiM achieves state-of-the-art performances.  
\end{itemize}

\section{Related Work}

\subsection{Contrastive Learning}
Contrastive learning approaches directly regularize the representation space by pushing representations of different views of the same image closer and spreading representations of views from different images apart. Contrastive learning requires randomly augmented views of source data to construct new data pairs. In an unsupervised manner, some methods treat different views from the same source as positive pairs, and views from different sources as negative pairs \cite{chen2020simple}. In a supervised way, with label supervision, views from the same class will be seen as positive pairs, and views from different classes will be regarded as negative pairs \cite{khosla2020supervised}. It is non-trivial to apply contrastive learning. First, stochastic augmentation for different views of samples is necessary and crucial to the performance.
Second, trivial solutions of the optimization problem should be avoided through using large batches of negative samples, momentum encoder \cite{he2020momentum, grill2020bootstrap} or stop-gradient scheme \cite{chen2020exploring}. 
\subsection{Semi-Supervised Learning}
Semi-supervised learning tries to utilize unlabeled data via self-training to achieve better performance. Typical semi-supervised learning methods perform self-training by pseudo-labeling unlabeled data and design extra regularization objectives. Two classes of regularization are mainly pursued and proved to be useful: consistency regularization \cite{tarvainen2017mean} and entropy minimization \cite{grandvalet2004semi}. The former encourages the model to generate consistent predictions on source data and randomly augmented views. The latter guides the model to output low-entropy predictions with confidence. Recently, MixMatch \cite{berthelot2019mixmatch} incorporates MixUp augmentations \cite{zhang2017mixup} and proposes a unified framework containing both of these regularizations. Following its success, UDA \cite{xie2020unsupervised}, ReMixMatch \cite{berthelot2019remixmatch} and FixMatch \cite{sohn2020fixmatch} proposes to use weakly augmented images to produce labels and enforce consistent predictions against strongly augmented samples through different designs. 
\subsection{Learning with Noisy Labels}
Many studies focus on reducing the effect of noise and generalizing from the correct label. On one hand, some methods explore ways to apply loss correction by estimating noise transition matrix \cite{patrini2017making,goldberger2016training}, re-weighting samples by designing criterions such as small-loss \cite{jiang2018mentornet,han2018co} and prediction disagreement \cite{malach2017decoupling}, or directly applying regularization through early-stop strategy \cite{liu2020early}. On the other hand, some methods focus on correcting wrong labels by learning class prototypes \cite{han2019deep}, predicting pseudo labels, or treating labels as learnable latent variables \cite{tanaka2018joint,yi2019probabilistic}.
Recently, DivideMix \cite{li2020dividemix} proposes to learn with noisy labels in a semi-supervised learning manner and achieves impressive performance. It detects the noisy samples by fitting a Gaussian Mixture Model (GMM) with the training loss, regards them as unlabeled samples, and applies modified MixMatch. DM-AugDesc \cite{nishi2021augmentation} further explores augmentation strategies to boost DivideMix. Also, some approaches attempt to leverage self-supervised pre-trained representation encoder through contrastive learning. REED \cite{zhang2020decoupling} tries to use it as the initial label corrector. C2D \cite{zheltonozhskii2021contrast} evaluates its effectiveness to initialize the model for different LNL methods such as DivideMix and ELR+ \cite{liu2020early}.


\section{Method}

\subsection{Empowering Semi-Supervised Learning with Contrastive Learning}
We first introduce general contrastive learning and semi-supervised learning algorithms and then propose CSSL, which combines the advantages of CL and SSL together.
\subsubsection{Contrastive Learning (CL)}
We introduce two algorithms here. Both self-supervised contrastive learning (SelfCon) and supervised contrastive learning (SupCon) algorithms have a stochastic augmentation function $A_{cntr}(.)$ and train a model $M_{CL} =\{Proj(F_1(.))|\theta_{feat,CL},\theta_{proj}\}$ with a feature extractor $F_1(.)$ parameterized by $\theta_{feat,CL}$ and a projector $Proj(.)$ parameterized by $\theta_{proj}$. WLOG, for a sample $x$, $\tilde{x} = A_{cntr}(x)$ is a randomly augmented view of $x$, the feature extractor will map $x$ to a representation vector $r=F_1(x) \in \mathcal{R}^{D_r}$, where $D_r$ is the dimension of representation space, and the projector will map $r$ to a vector $z=Proj(r) \in R^{D_z}$. Given a bunch of $K$ pairs of data with label $\{\textbf{d}_k, \textbf{l}_k\}_{k=1}^{K}$, both algorithms need $2K$ augmented pairs $\{\tilde{\textbf{d}}_b, \tilde{\textbf{l}}_b\}_{b=1}^{2K}$ for training, where $\tilde{\textbf{d}}_{2k}$ and $\tilde{\textbf{d}}_{2k-1}$ are two different augmented views of $\textbf{d}_k$ through $A_{cntr}(.)$ and $\tilde{\textbf{l}}_{2k}=\tilde{\textbf{l}}_{2k-1}=\textbf{l}_k$($k=1...K$). In other words, two views are generated for each data source. The family of contrastive loss basically follows Info-NCE loss \cite{oord2018representation}, which tries to maximize/minimize the mutual information of positive/negative pairs. The main difference between SelfCon and SupCon happens during loss calculation, as SelfCon will not use label supervision while SupCon will take categories into consideration. Let $i\in I \triangleq \{1,2,..,2K\}$, $C(i) \triangleq I \backslash \{i\}$, and $j(i)$ be the index of the other augmented view from the same source. Given the projected vector $\{z_k\}_{k=1}^{2K}$, SelfCon calculates the following loss:
\begin{equation}
    \begin{aligned}
    \mathcal{L}_{self} = - \sum_{i \in I} \log \frac{exp(z_i\cdot z_{j(i)}/\tau)}{\sum_{c \in C(i)}exp(z_i\cdot z_c/\tau)}
    \end{aligned}
\end{equation}
Here $\tau$ is the temperature hyperparameter. The numerator counts for positive pairs, and the denominator contains both positive and negative pairs. Let $S(i) \triangleq \{s \in C(i): \tilde{\textbf{l}}_s=\tilde{\textbf{l}}_i\}$, SupCon calculates the following loss:
\begin{equation}
    \begin{aligned}
    \mathcal{L}_{sup} = \sum_{i \in I} \frac{-1}{|S(i)|}\sum_{s \in S(i)}\log \frac{exp(z_i\cdot z_{s}/\tau)}{\sum_{c \in C(i)}exp(z_i\cdot z_c/\tau)}
    \end{aligned}
\end{equation}
For each anchor vector $z_i$, only the other view generated from the same source $z_{j(i)}$ is seen as positive in SelfCon, yet in SupCon all the other views generated from data with the same label are seen as positive.
\subsubsection{Semi-Supervised Learning (SSL)}
Consider a partially-labeled dataset $\mathcal{D}=\{\mathcal{X}, \mathcal{U}\}$, $\mathcal{X}=\{(x_i,y_i)\}_{i=1}^{N}$ and $\mathcal{U}=\{u_j\}_{j=1}^{M}$, where $\{x_i,u_j\}$ are samples and $y_i \in \{0,1\}^C$ is the one-hot label vector over $C$ classes. Semi-supervised learning algorithms solve a $C$-class classification task by training a model $M_{SSL}=\{G(F_2(.))|\theta_{feat,SSL}, \theta_{cls}\}$ with a feature extractor $F_2(.)$ parameterized by $\theta_{feat,SSL}$ and a cascaded classifier $G(.)$, parameterized by $\theta_{cls}$. Many successful SSL algorithms try to exploit unlabeled data with consistency regularization, entropy minimization, and randomized augmentation. Specifically, let $Semi\_Alg(\mathcal{X}, \mathcal{U}, \mathcal{H}, \mathcal{F})$ be the chosen semi-supervised learning algorithm, where $\mathcal{H}$ and $\mathcal{F}$ are the set of hyperparameters and functions. For each training epoch, it tries to generate an augmented labeled set $\mathcal{X'}=\{x_i',p_i'\}_{i=1,2,...}$ and an unlabeled set $\mathcal{U'}=\{u_j', q_j'\}_{j=1,2,...}$, here $p_i, q_j$ refers to processed labels. Let $A_{semi}(.)$ be a stochastic augmentation function. Then, it will minimize the following objectives:
\begin{equation}
\begin{aligned}
    &\mathcal{L}_{Semi} = \overbrace{\frac{1}{|\mathcal{X}'|}\sum_{x,p \in \mathcal{X}'}H(p, p_{M_{SSL}}(A_{semi}(x)))}^{\mathcal{L}_x}
    + \\ &\lambda_u\cdot\overbrace{\frac{1}{|\mathcal{U}'|}\sum_{u,q \in \mathcal{U}'}H(q, p_{M_{SSL}}(A_{semi}(u)))}^{\mathcal{L}_u} + \lambda_{r}\mathcal{L}_{reg}.   
\end{aligned}
\end{equation}
Note, to measure entropy between processed labels and model's predictions, Cross-Entropy (CE) loss and L2-Loss are commonly uesd. 
\subsubsection{Contrastive Semi-Supervised Learning (CSSL)}
Now we study how to combine CL and SSL into one unified algorithm. We propose a general multi-task learning algorithm (Alg.\ref{alg:ALG1}) that employs SupCon to utilize label supervision of the labeled set, and uses SelfCon in two ways: 1) to provide self-supervised representation learning (a.k.a. SelfCon pre-training) on the whole dataset before multi-task learning; 2) to keep learning self-supervised features from the unlabeled set during the multi-objective optimization. Because the optimization objectives of CL and SSL are different, we use a model $M_{CSSL}=$ $\{G(F(.)), Proj(F(.))|\theta_{feat}, \theta_{cls}, \theta_{proj}\}$ with two different heads $G(.)$ and $Proj(.)$, and one feature extractor $F(.)=F_1(.)=F_2(.)$, by sharing the weights of $\theta_{feat,SSL}$ and $\theta_{feat,CL}$. For the sake of generality, we slightly abuse the notation of $Semi\_Alg$ and wrap up hyperparameters and functions used for SSL algorithm with $\mathcal{H}, \mathcal{F}$ (e.g. $\lambda_{u}, \lambda_{r} \in \mathcal{H}$ and $A_{semi}(.) \in \mathcal{F}$). Thus, one advantage of CSSL is that many popular SSL algorithms (e.g. Mixmatch, ReMixMatch, and Fixmatch) can be directly plugged in and contributed as the SSL module without inner modification at all. We apply SupCon/SelfCon to data batches from labeled/unlabeled set to match the style of SSL algorithm.
\begin{algorithm}
\renewcommand{\algorithmicrequire}{\textbf{Input:}}
\renewcommand{\algorithmicensure}{\textbf{Output:}}
\renewcommand{\algorithmiccomment}{ \ \ \ // }
\caption{A Multi-task Contrastive Semi-Supervised Learning Algorithm with Pre-training. }
\begin{algorithmic}[1]
\REQUIRE Dataset $\mathcal{D}=\{\mathcal{X}, \mathcal{U}\}$, pre-training and multi-task training steps $N_1, N_2$, Supcon and SelfCon loss weight $\lambda_{sup}, \lambda_{self}$, temperature $\tau_1,\tau_2,\tau_3$, SSL used hyperParamer set $\mathcal{H}$, SSL used function set $\mathcal{F}$
\WHILE{$t<N_1$} \COMMENT{SelfCon pre-training}
    \STATE Draw raw data batch $\hat{D}_B$ from $\mathcal{D}$ \COMMENT{Ignore labels} 
    \STATE $\hat{D}_{B,1}=A_{cntr}(\hat{D}_B)$  \COMMENT{Different views} 
    \STATE $\hat{D}_{B,2}=A_{cntr}(\hat{D}_B)$  \COMMENT{As $A_{cntr}(.)$ is stochastic} 
    \STATE $\mathcal{L}_{self}= SelfCon(\hat{D}_{B,1},\hat{D}_{B,2}, \tau_1)$
    \STATE $\theta_{feat}, \theta_{proj}=\text{SGD}(\mathcal{L}_{self}, \theta_{feat}, \theta_{proj})$
\ENDWHILE
\WHILE{$t<N_2$} \COMMENT{Multi-task learning}
    \STATE Draw data batch $\{\hat{X}_B, \hat{Y}_B\}, \hat{U}_B$ from $\mathcal{X}, \mathcal{U}$ 
    \FOR{$j=1,2$}
    \STATE $\hat{X}_{B,j}=A_{cntr}(\hat{X}_B)$ 
    \STATE $\hat{U}_{B,j}=A_{cntr}(\hat{U}_B)$ 
    \ENDFOR
    \STATE $\mathcal{L}_{self}= SelfCon(\hat{U}_{B,1},\hat{U}_{B,2},\tau_2)$
    \STATE $\mathcal{L}_{sup}= SupCon(\hat{X}_{B,1},\hat{X}_{B,2}, \hat{Y}_B, \tau_3)$
    \STATE $\mathcal{L}_{Semi}, \hat{X}'_B, \hat{U}'_B = Semi\_Alg(\hat{X}_B, \hat{U}_B, \hat{Y}_B, \mathcal{H},\mathcal{F})$ \COMMENT{$\hat{X}'_B, \hat{U}'_B$ are intermediate augmented data}
    \STATE $\mathcal{L}= \mathcal{L}_{semi} + \lambda_{sup}\mathcal{L}_{sup} + \lambda_{self} \mathcal{L}_{self}$
    \STATE $\theta_{feat}, \theta_{proj}, \theta_{cls}=\text{SGD}(\mathcal{L}, \theta_{feat}, \theta_{proj},\theta_{cls})$
\ENDWHILE
\end{algorithmic}
\label{alg:ALG1}
\end{algorithm}
\subsubsection{CL Introduces Extra Consistency Regularization }
A key factor to the success of semi-supervised learning methods is to pursue consistency regularization, which encourages the model outputs same predictions for input with small perturbation. Concretely, recall the formula that calculates $\mathcal{L}_{Semi}$, we can see consistency regularization has been implicitly enforced as for a processed sample $x$ or $u$, the algorithm will try to minimize the entropy between the processed labels and predictions of its augmented views, i.e. $A_{semi}(x)$ or $A_{semi}(u)$. Recently, a study \cite{wei2020theoretical} proposes a unified theoretical analysis on this kind of self-training with constructed consistency regularization, by assuming expansion effect. Specifically, let $P_i$ be the data distribution conditioned on class label $i$. For a small subset $S$ of samples labeled $i$, expansion effect assumes that, 
\begin{equation}
    \begin{aligned}
    P_i(\text{neighbourhood of }S) \geq cP_i(S) 
    \end{aligned}
\end{equation}
Here, $c > 1$ is the expansion factor, and the neighbourhood of $S$ is defined to introduce data augmentation. Generally speaking, the neighbourhood of $S$ can be sampled by applying a stochastic augmentation function to samples in $S$. This expansion assumption indicates that data distribution within each class has good continuity. With this assumption, consistency regularization can be defined as:
\begin{equation}
    \begin{aligned}
    \mathcal{R}(G(F)) = \mathbb{E}_x\mathop{max}\limits_{neighbor \ x'} \mathbf{1}(G(F(x'))\neq G(F(x)))
    \end{aligned}
\end{equation}
\begin{figure}[t]
\centering
\includegraphics[width=0.8\linewidth]{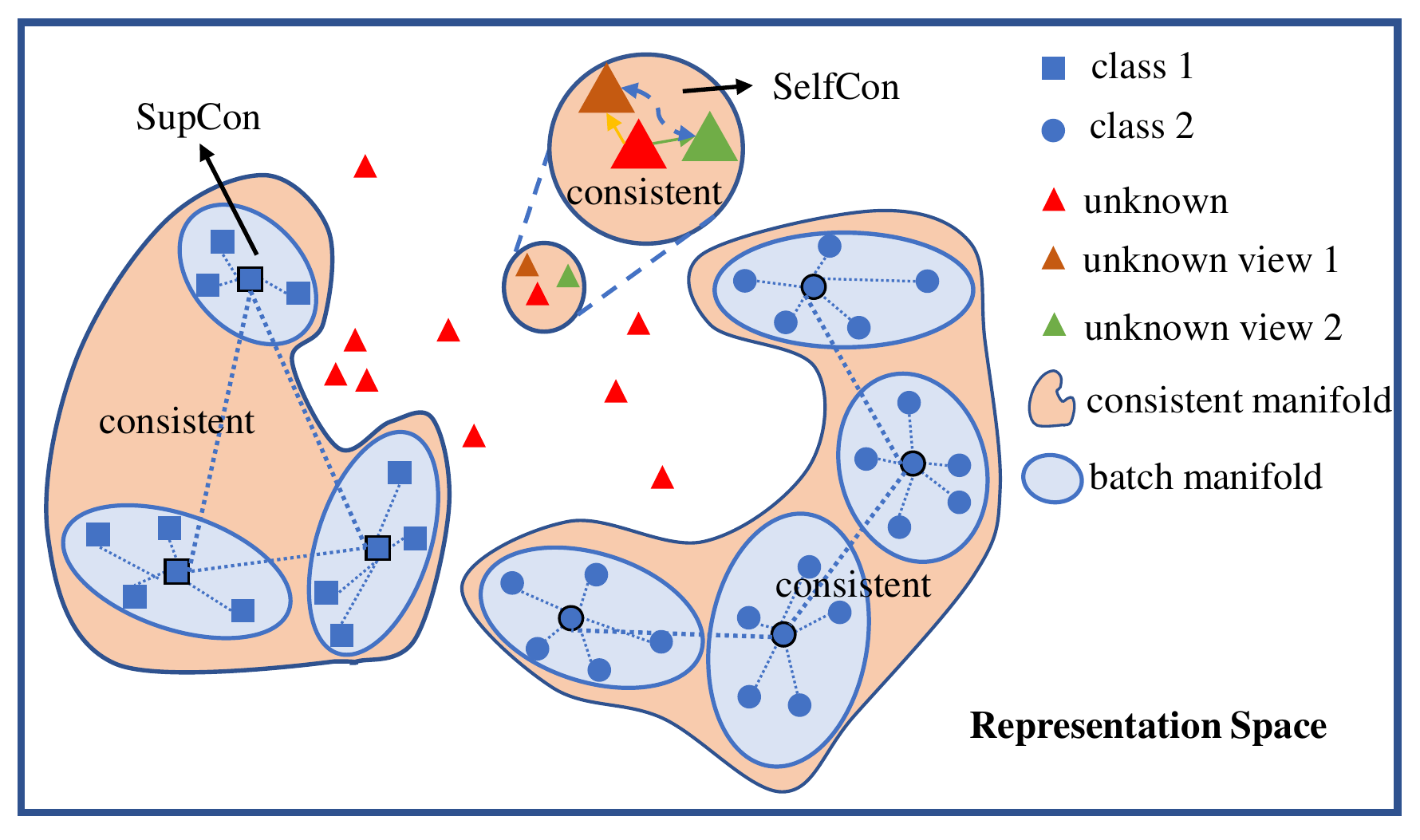}
\caption{An example on how contrastive learning introduces extra consistency regularization. Models maintain consistent predictions for representations in orange manifolds. Both SelfCon and SupCon encourage representation space to maintain consistency in the batch. Given data batches drew from dataset, SelfCon only learn a small consistent manifold around each sample based on `self-supervision'. SupCon can learn a more continuous manifold as samples with same label can be clustered together.}
\label{fig2}
\end{figure}
Contrastive learning methods also use randomized data augmentation techniques to produce `weak supervision'. Researches have empirically provided positive evidences that representation encoder can also benefit from such supervision even if using a non-linear MLP $Proj_{\theta_{proj}}(.)$. We suggest that this kind of `weak supervision' also implicitly implies consistency regularization to the representation space by regularizing the weights of the feature extractor(as illustrated in Fig.\ref{fig2}). On one hand, SelfCon builds `self-supervision' for different views from the same sample. Recalling the SelfCon loss, for a sample $d$, suppose $\hat{d}_1, \hat{d}_2$ are two different augmented views of $d$ and let $\hat{z}_1= Proj_{\theta_{proj}}(F_{\theta_{feat}}(\hat{d}_1))$, $\hat{z}_2= Proj_{\theta_{proj}}(F_{\theta_{feat}}(\hat{d}_2))$, an easy way to minimize the distance between $\hat{z_1}$ and $\hat{z}_2$ is to encourage the feature extractor learn to map a sample and its neighbour in data space to similar representation, i.e, $\hat{r}_1 =F_{\theta_{feat}}(\hat{d}_1)$ and $\hat{r}_2=F_{\theta_{feat}}(\hat{d}_2)$ should be similar. On the other hand, SupCon further tries to cluster data from the same class in the projected space, which further encourages the feature extractor to learn a more continuous representation conditioned on class label. This empirically lead the model to better fit expansion property and have more consistent predictions on augmented samples. 
\subsection{CSSL with Noisy Labels}
We adapt CSSL to solve LNL tasks by first introducing some key designs to leverage SSL algorithms, and then propose CoDiM for LNL tasks with several simple yet critical modifications inspired by DivideMix.
\begin{algorithm}[!h]
\renewcommand{\algorithmicrequire}{\textbf{Input:}}
\renewcommand{\algorithmicensure}{\textbf{Output:}}
\renewcommand{\algorithmiccomment}{ \ \ \ // }
\caption{CoDiM: A Learning with Noisy Labels Algorithm via Contrastive Semi-Supervised Learning. }
\begin{algorithmic}[1]
\REQUIRE Dataset $\mathcal{D}=\{\mathcal{X}, \mathcal{Y}\}$, SelfCon, SupCon, Warmup, training steps $N_1, N_2$, epoch $E$, Contrastive loss weight $\lambda_{cl}$, temperature $\tau_1,\tau_2$, SimCLR and strong augmentation function $A_{simc}(.)$, $A_{sa}(.)$, D\_MixMat hyperParamer set $\mathcal{H}$, D\_MixMat function set $\mathcal{F}=\{A_{wa}(.), A_{sa}(.),\dots\}$, algorithm mode \textit{mode}
\WHILE{$t<N_1$} \Comment{SelfCon pre-training}
    \STATE Draw raw data batch $\hat{D}_B$ from $\mathcal{D}$ \Comment{Ignore labels}
    \STATE $\hat{D}_{B,1}=A_{simc}(\hat{D}_B)$  \Comment{Different views}
    \STATE $\hat{D}_{B,2}=A_{simc}(\hat{D}_B)$  
    \STATE $\mathcal{L}_{self}= SelfCon(\hat{D}_{B,1},\hat{D}_{B,2}, \tau_1)$
    \STATE $\theta_{feat}, \theta_{proj}=\text{SGD}(\mathcal{L}_{self}, \theta_{feat}, \theta_{proj})$
\ENDWHILE
\STATE $\theta^{(1)} = \{\theta_{feat}^{(1)}, \theta_{proj}^{(1)}, \theta_{cls}^{(1)}\} = \{\theta_{feat}, \theta_{proj}, \tilde{r}(\theta_{cls})\}$
\STATE $\theta^{(2)} = \{\theta_{feat}^{(2)}, \theta_{proj}^{(2)}, \theta_{cls}^{(2)}\} = \{\theta_{feat}, \theta_{proj}, \tilde{r}(\theta_{cls})\}$
\Comment{Randomly initialized $\theta_{cls}$}
\STATE $\theta^{(1)}, \theta^{(2)}=$WarmUp$(\mathcal{D}, \theta^{(1)}, \theta^{(2)},N_2)$ \Comment{Initialize \& WarmUp classifier}
\WHILE{$e<E$} \Comment{Learning with label noise}
    \STATE $\mathcal{C}^{(1)}, \mathcal{I}^{(1)}=$GMM$(\mathcal{D}, \theta^{(2)})$
    \STATE $\mathcal{C}^{(2)}, \mathcal{I}^{(2)}=$GMM$(\mathcal{D}, \theta^{(1)})$
    \WHILE{iter$\ < \ $num\_iter}
    \FOR{$j=1,2$}
    \STATE Draw data batch $\{\hat{X}_B, \hat{Y}_B\}$ from $\mathcal{C}^{(j)}$  
    \STATE Draw data batch $\{\hat{U}_B\}$ from $\mathcal{I}^{(j)}$  
    \STATE $\hat{X}_{B,1}=A_{sa}(\hat{X}_B)$, $\hat{X}_{B,2}=A_{sa}(\hat{X}_B)$
    \STATE \textbf{if} \textit{mode} is SelfCon \textbf{do}
    \STATE \ \ \ \ $\mathcal{L}_{cl}= SelfCon(\hat{X}_{B,1},\hat{X}_{B,2}, \tau_2)$
    \STATE \textbf{else if} \textit{mode} is SupCon \textbf{do}
    \STATE \ \ \ \ $\mathcal{L}_{cl}= SupCon(\hat{X}_{B,1},\hat{X}_{B,2}, \hat{Y}_B, \tau_2)$
    \STATE \textbf{end if}
    \STATE $\mathcal{L}_{Semi}, \hat{X}'_B, \hat{U}'_B= $D\_MixMat$(\hat{X}_B, \hat{U}_B, \mathcal{H}, \mathcal{F})$
    \STATE \Comment{$D\_MixMat$} refers to the used SSL method
    \STATE $\mathcal{L}= \mathcal{L}_{semi} + \lambda_{cl}\mathcal{L}_{cl}$ 
    \STATE $\theta^{(j)}=\text{SGD}(\mathcal{L}, \theta^{(j)})$    
    \ENDFOR
    \ENDWHILE
\ENDWHILE
\end{algorithmic}
\label{alg:ALG2}
\end{algorithm}
\subsubsection{Learning with Noisy Labels via SSL}
Consider a set of training data with label noise $\tilde{\mathcal{D}}= \{(\tilde{x}_{i}, \tilde{y}_{i})\}_{i=1}^{N}$ where $\tilde{y}_i \in \{0,1\}^{\tilde{C}}$, our target is to learn from correct label supervision as well as avoid over-fitting incorrect labels without any prior knowledge of noise distribution. Here we only consider the case that no extra clean labeled data are accessible, and the only guarantee is that for each class $i$, the population of samples with correct labels are larger than any other population of samples that should have correct labels $j \in \{1,...,\tilde{C}\} \backslash \{i\}$ but are incorrectly labeled with $i$. Let $\mathcal{C}, \mathcal{I}$ be the set of samples in $\mathcal{\tilde{D}}$ that are correctly/incorrectly labeled, if knowing the partition of $\mathcal{C}$ and $\mathcal{I}$, one can solve it as a semi-supervised learning task. However, it's challenging to specify the partition. A direct extension is to design a two-stage algorithm which tries to decide a partition that divides $\mathcal{\tilde{D}}$ into $\mathcal{\tilde{C}}$ and $\mathcal{\tilde{I}}$ via a noise detection module, and then apply an SSL algorithm on $\{\mathcal{\tilde{C}}, \mathcal{\tilde{I}}\}$. Since the noise detection task can be regarded as a clean-or-noisy classification task, the noise detection module typically has a module which measures the dataset and decide a partition thershold $\mathcal{\tilde{R}}(\mathcal{T})$. A commonly-used measure is to choose samples with lower training loss based on the SSL classifier. To better leverage this measure, warming-up the classifier by training with traditional CE-loss for a few epochs is also a good choice.
\subsubsection{Adapting CSSL to CoDiM}
Generally speaking, CoDiM also serves as a two-phase algorithm, which is specified in Alg.\ref{alg:ALG2} (for brevity, details of SSL are summarized as $\mathcal{H}$ and $\mathcal{F}$). In the first phase, SelfCon pre-training will be applied using all data ignoring labels, and then a very short `warming up' using CE-loss will be used as the initialization of the classifier head. In the second phase, CoDiM first decides the partition imitating DivideMix via fitting a GMM model to choose samples with lower classification loss as clean samples. Then it will apply contrastive semi-supervised learning based on the partition, taking the modified MixMatch algorithm used in DivideMix as the SSL module. However, one critical change is that here CoDiM only apply SupCon or SelfCon to the possibly clean set as we find that keep applying SelfCon to the possibly noisy set will downgrade the performance. Also, when dealing with high ratio label noise or noise among similar classes, we suggest to replace SupCon with SelfCon to learn from possibly clean set to further avoid learning from biases. We follow DivideMix to use `co-divide', which uses two networks, and for each iteration, one network use the partition threshold decided by the other. Also, we find that other customized techniques proposed by DivideMix (e.g. label co-guessing and co-refinement) can be maintained here. Following the `AugDesc-WS' augmentation strategy \cite{nishi2021augmentation}, we use so-called `weak augmentation' (random crop and flip) to generate views for querying prediction, use so-called `strong augmentation' (AutoAugment) to generate views for gradient descent. We further note in pre-training phase, we use augmentation functions proposed by SimCLR \cite{chen2020simple} in SelfCon to get better pre-training results, and use same `strong augmentation' used in `AugDesc-WS' in SupCon/SelfCon during the second phase to reduce the computation cost and in some sense reduce the difficulty of the optimization problem, which are both critical.
\section{Experiments}
\subsection{Dataset and Experimental Setup}
We conduct multiple experiments on CIFAR-10 and CIFAR-100 \cite{krizhevsky2009learning} for SSL tasks and LNL problems. The two datasets both contain $50K$ training and $10K$ test images of size $32 \times 32$ from 10 and 100 classes, respectively. Following previous work \cite{li2020dividemix}, we use PreAct Resnet18 as the feature extractor. We first examine the performance of our CSSL algorithm on CIFAR-10 under two ratios of labeled samples (20\% and 80\%, respectively). We then evaluate CoDiM on learning with different types and levels of synthetic label noise. Two types of label noise: \textit{symmetric} and \textit{asymmetric} are tested. Symmetric noise is produced by selecting a percentage of the training data and assigning them uniformly random labels. Asymmetric noise is generated to simulate real world noise, where only the labels of similar classes will be assigned. We then apply CoDiM on ANIMAL-10N \cite{song2019selfie} and WebVision \cite{li2017webvision}, two datasets with real world label noise. ANIMAL-10N contains 5 pairs of confusing animals with $55K$ noisy human-labeled online images in total. The noisy label ratio is about 8\%. We use VGG19 backbone to stay consistent with previous work. WebVision contains $2.4M$ images collected by searching the 1,000 concepts in ImageNet ILSVRC12 on the Internet. For fair comparison, we use the inception-resnet v2 to evaluate the first 50 classes of the Google image subset. More implementation details are described in Appendix A.
\begin{table}[h]
\begin{center}
\begin{tabular}{lcccccccccc}
\hline
Dataset & & \multicolumn{2}{c}{CIFAR-10} \\
\hline
Methods/labeled ratio & & 20\% & 80\%  \\
\hline
\textbf{w/o SelfCon pre-training} \\
\hline
\multirow{2}{*}{SSL \textbf{(Fig 3.a)}} & Best & 89.3 & 96.2 \\
& Last & 89.0 & 96.0 \\
\hline
\multirow{2}{*}{SSL-L(Self)-U(Self)} & Best & 89.1 & 96.2 \\
& Last & 88.9 & 96.0 \\
\hline
SSL-L(Sup)-U(Self) & Best & 91.7 & 96.6 \\
\textbf{(Fig 3.b)}& Last & 91.5 & 96.4 \\
\hline
\textbf{w/ SelfCon pre-training} \\
\hline
\multirow{2}{*}{SSL \textbf{(Fig 3.c)}} & Best & 94.0 & 96.8 \\
& Last & 93.9 & 96.7 \\
\hline
\multirow{2}{*}{SSL-L(Self)-U(Self)} & Best & 94.6 & 96.6 \\
& Last & 94.4 & 96.5 \\
\hline
SSL-L(Sup)-U(Self) & Best & \textbf{94.7} & \textbf{96.9} \\
\textbf{(Fig 3.d)} \textbf{(CSSL)}& Last & \textbf{94.4} & \textbf{96.8} \\
\hline
\end{tabular}
\end{center}
\caption{Comparing CSSL with basic SSL algorithm in terms of test accuracy. \textit{S}(\textit{alg}) denotes applying \textit{alg} on samples drew from set \textit{S} in the second phase. \textit{S} $\in$\{\textbf{L}abeled,\textbf{U}nlabeled\}, \textit{alg} $\in$ \{\textbf{Self}Con, \textbf{Sup}Con\}. }
\label{tab:table1}
\end{table}
\subsection{SelfCon Pre-training and Contrastive Metrics Improves Performance of SSL}
Note that though being an algorithm that is compatible with many SSL algorithms, here we only evaluate a certain realization of CSSL, which uses a modified version of MixMatch used in DivideMix as the SSL module, as such algorithm will also be used by CoDiM. We report the results in Table~\ref{tab:table1} and address two key observations. First, SelfCon pre-training improves the performance of SSL, especially when the labeled ratio is low, as accuracy of all methods with SelfCon pre-training boost 0.3\%-0.6\% given 80\% label and 3\%-5\% given 20\% label. This also supports the discovery that SelfCon pre-training provides more robust results when dealing with high ratio label noise. Secondly, contrastive learning helps the performance of the classifier, as CSSL always outperforms basic SSL algorithms in both cases. This empirically supports our claim that contrastive learning will further provide consistency regularization. We also show that methods that leverage contrastive learning tend to have more clustered representations via showing t-SNE visualization of data representations of test set, certain experimental cases. Besides comparing the performance of CSSL and modified MixMatch on CIFAR-10, we also apply ablation studies to show the effect of each extension contained in CSSL, which can be found in Appendix B.
\begin{figure}[htbp]
\centering
\subfigure[SSL w/o pre-training]{
\includegraphics[width=0.4\linewidth]{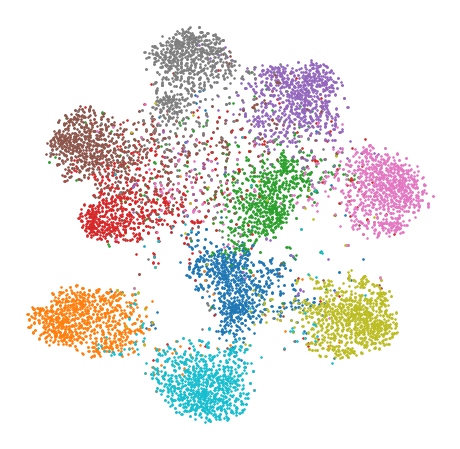}
}
\quad
\subfigure[CSSL w/o pre-training]{
\includegraphics[width=0.4\linewidth]{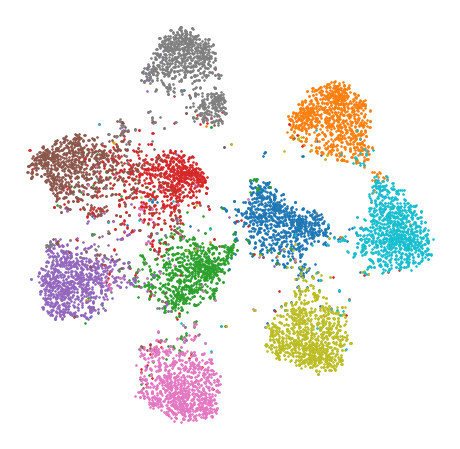}
}
\quad
\subfigure[SSL w/ pre-training]{
\includegraphics[width=0.4\linewidth]{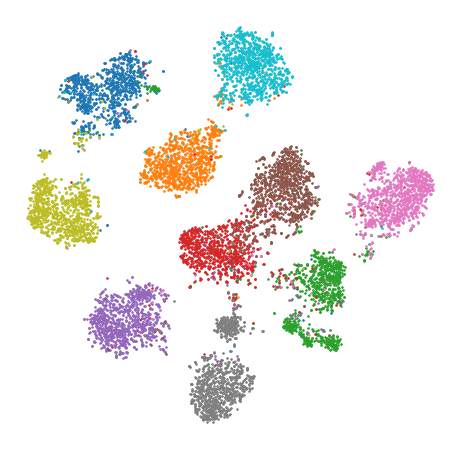}
}
\quad
\subfigure[CSSL w/ pre-training]{
\includegraphics[width=0.4\linewidth]{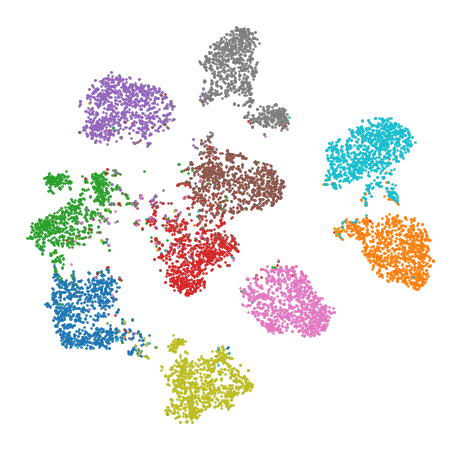}
}
\caption{t-SNE visualizations of different methods on the test set of CIFAR-10. `Pre-training' refers to pre-train a model using SelfCon. Comparing horizontally, it is obvious that leveraging SelfCon pre-training largely promotes the clustering of each class. Comparing vertically, we can see that using SupCon regularization further sets clusters apart. }
\end{figure}
\subsection{CoDiM on Noisy CIFAR-10 and CIFAR-100}
We compare two versions of CoDiM, depending on which contrastive learning algorithm are used in the second phase, namely CoDiM-Sup(use SupCon) and CoDiM-Self(use SelfCon) with other baseline methods on CIFAR-10 and CIFAR-100 with different levels and types of label noise. Results gained by proposed methods, important baselines, and two ablation studies are shown in Table~\ref{tab:table2}. We see that in all cases CoDiM achieves state-of-the-art performances. In the symmetric case, while the noise ratio is not extremely high, CoDiM-Sup outperforms other methods. However, CoDiM-Self shows competitive performances under high ratios of symmetric and asymmetric noise. This suggests that CoDiM combines online contrastive learning with semi-supervised learning in a simple yet better way. More ablation studies and visualizations results are in Appendix C and F.   
\begin{table*}[h]
\begin{center}
\begin{tabular}{lcccccccccc}
\hline
Dataset&  & \multicolumn{5}{c}{CIFAR-10} & \multicolumn{4}{c}{CIFAR-100} \\
\hline
Noise type & & \multicolumn{4}{c}{Sym.}& Asym.  & \multicolumn{4}{c}{Sym.} \\
\hline
Method/Noise ratio & & 20\% & 50\% & 80\% & 90\% & 40\% & 20\% & 50\% & 80\% & 90\% \\
\hline 
\multirow{2}{*}{Cross-Entropy} & Best & 86.8 & 79.4 & 62.9 & 42.7 & 85.0 & 62.0 & 46.7 & 19.9 & 10.1 \\
 & Last & 82.7 & 57.9 & 26.1 & 16.8 & 72.3 & 61.8 & 37.3 & 8.8 & 3.5 \\
\multirow{2}{*}{M-correction \cite{arazo2019unsupervised}} & Best & 94.0 & 92.0 & 86.8 & 69.1 & 87.4 & 73.9 & 66.1 & 48.2 & 24.3 \\
 & Last & 93.8 & 91.9 & 86.6 & 68.7 & 86.3 & 73.4 & 65.4 & 47.6 & 20.5 \\
\multirow{2}{*}{DivideMix \cite{li2020dividemix}} & Best & 96.1 & 94.6 & 93.2 & 76.0 & 93.4 & 77.3 & 74.6 & 60.2 & 31.5 \\
 & Last & 95.7 & 94.4 & 92.9 & 75.4 & 92.1 & 76.9 & 74.2 & 59.6 & 31.0 \\
\multirow{2}{*}{DM-AugDesc \cite{nishi2021augmentation}} & Best & 96.3 & 95.4 & 93.8 & 91.9 & 94.6 & 79.5 & 77.2 & 66.4 & 41.2 \\
 & Last & 96.2 & 95.1 & 93.6 & 91.8 & 94.3 & 79.2 & 77.0 & 66.1 & 40.9 \\
\multirow{2}{*}{C2D \cite{zheltonozhskii2021contrast}} & Best & 96.4 & 95.5 & 94.3 & 93.2 & 93.6 & 78.7 & 76.8 & 59.8 & 52.1 \\
 & Last & 96.3 & 95.3 & 94.2 & 93.0 & 93.3 & 78.4 & 76.4 & 59.6 & 51.9 \\
\multirow{2}{*}{REED \cite{zhang2020decoupling}} & Best & 95.9 & 95.4 & 94.4 & \underline{93.6} & 92.4 & 76.9 & 72.8 & 65.6 & \underline{55.7} \\
 & Last & 95.8 & 95.2 & 94.2 & \underline{93.5} & 92.3 & 76.7 & 72.5 & 65.4 & \underline{55.2} \\
\hline
CoDiM-bare & Best & 96.4 & 95.5 & \underline{94.6} & 93.4 & 94.4 & 80.6 & 77.5 & 60.5 & 52.6 \\
(only pre-training) & Last & 96.3 & 95.3 & \underline{94.5} & 93.3 & 94.1 & 80.4 & 77.2 & 60.2 & 52.3 \\
\hline
CoDiM-CSSL & Best & \underline{96.5} & \underline{96.1} & 94.6 & 93.5 & 94.2 & \underline{80.7} & \underline{78.0} & \underline{63.6} & 54.5 \\
(add SelfCon on $\mathcal{I}$)& Last & \underline{96.3} & \underline{96.0} & 94.4 & 93.4 & 94.0 & \underline{80.5} & \underline{77.8} & \underline{63.4} & 54.3 \\
\hline
\hline
\multirow{2}{*}{\textbf{CoDiM-Self (Ours)}} & Best & \underline{96.5} & 95.3 & 94.3 & 92.9 & \textbf{95.5} & 80.4 & 77.2 & \underline{63.6} & \textbf{56.4} \\
& Last & \underline{96.3} & 95.2 & 94.2 & 92.7 & \textbf{95.3} & 80.2 & 77.1 & \underline{63.4} & \textbf{56.1} \\
\hline
\multirow{2}{*}{\textbf{CoDiM-Sup (Ours)}} & Best & \textbf{97.0} & \textbf{96.5} & \textbf{94.7} & \textbf{93.7} & \underline{95.2} & \textbf{81.1} & \textbf{78.1} & \textbf{67.3} & 55.2 \\
 & Last & \textbf{96.9} & \textbf{96.4} & \textbf{94.6} & \textbf{93.4} & \underline{95.0} & \textbf{80.9} & \textbf{77.9} & \textbf{67.0} & 54.9 \\
\hline
\end{tabular}
\end{center}
\caption{Comparison with existing methods on CIFAR-10/100 with different noise settings. We re-implement C2D and REED here. Note CoDiM-bare can be regarded as a combination of SelfCon pre-training and then apply DM-AugDesc. CoDiM-CSSL apply SupCon on $\mathcal{C}$ and SelfCon on $\mathcal{I}$ following CSSL. Full table of results can be found in Appendix D.}
\label{tab:table2}
\end{table*}
\begin{table}[h]
\begin{center}
\begin{tabular}{lc}
\hline
Method & Test Acc \\
\hline
SELFIE \cite{song2019selfie} & 81.8 \\
PLC \cite{zhang2020learning} & 83.4 \\
Nested Co-teaching \cite{chen2021boosting} & 84.1 \\
DivideMix (w/o ImageNet pre-training) & 85.8 \\
DM-AugDesc \cite{nishi2021augmentation}& 86.0 \\
DivideMix \cite{li2020dividemix} & 88.8\\
C2D \cite{zheltonozhskii2021contrast} & 88.9\\
DM-AugDesc \cite{nishi2021augmentation}& \underline{89.1} \\
\hline
CoDiM-bare & \underline{89.1} \\
CoDiM-Sup & \textbf{89.2} \\
CoDiM-Self & \textbf{89.4} \\
\hline
\end{tabular}
\end{center}
\caption{Comparison with existing methods on ANIMAL-10N. We re-implement methods starting from DivideMix.}
\label{tab:table3}
\end{table}
\subsection{CoDiM on Real World Noisy Datasets}
Table \ref{tab:table3} shows the results on the ANIMAL-10N dataset. All CoDiM-style methods achieve 89\%+ accuracy. These results clearly show the improvements gained by adding different components of CoDiM like SelfCon pre-training, augmentation strategy, and extra contrastive learning scheme, as C2D can be regarded as using SelfCon pre-training \& DivideMix and CoDiM-bare can be regarded as SelfCon pre-training \& DM-AugDesc. CoDiM-Self and CoDiM-Sup beat basic DivideMix and DM-AugDesc by large and even beat the updated versions with prior information learned from extra data domain(via using model pre-trained on ImageNet). Here CoDiM-Self achieves the state-of-the-art performance, and we conjecture this is due to the type of noise in ANIMAL-10N is closer to asymmetric noise. \\
Table \ref{tab:table4} shows the results on WebVision. The full version table can be found in Appendix E. We see that CoDiM-Sup achieves the best performance on the WebVision validation set, while CoDiM-Self has the best generalization performance on the ILSVRC12 validation set. As C2D has shown the advantages of applying SelfCon pre-training, our methods show the performance engagement on large-scale real-world noisy datasets by further providing contrastive regularization, designing suitable augmentation strategies, and combine all of these techniques in a harmonious way.
\begin{table}[h]
\begin{center}
\begin{tabular}{lcccc}
\hline
\multirow{2}{*}{Method} & \multicolumn{2}{c}{WebVision} & \multicolumn{2}{c}{ILSVRC12} \\
 & top1 & top5 & top1 & top5 \\
\hline
Co-teaching 
& 63.58 & 85.20 & 61.48 & 84.70 \\
Iterative-CV 
& 65.24 & 85.34 & 61.60 & 84.98 \\
DivideMix 
& 77.32 & 91.64 & 75.20 & 90.84 \\
ELR+ 
& 77.78 & 91.68 & 70.29 & 89.76 \\
LongReMix 
& 78.92 & 92.32 & - & - \\
DM-AugDesc 
& 78.64 & 93.20 & 75.52 & 92.12 \\ 
GJS 
& 79.28 & 91.22 & 75.50 & 91.27 \\
C2D 
& 80.20 & 93.22 & 76.64 & 92.32 \\ 
\hline
CoDiM-bare  & 80.32 & \underline{93.40} & 76.60 & 92.36 \\
CoDiM-Self & 80.12 & 93.52 & \textbf{77.24} & \underline{92.48}\\ 
CoDiM-Sup & \textbf{80.88} & 92.48 & 76.52 & 91.96 \\
\hline
\end{tabular}
\end{center}
\caption{Comparison with existing methods on WebVision. We re-implement DM-AugDesc and C2D. We provide references of some baseline methods here: Co-teaching \cite{han2018co}, Iterative-CV \cite{chen2019understanding}, LongReMix \cite{cordeiro2021longremix}, GJS \cite{englesson2021generalized}.}
\label{tab:table4}
\end{table}

\section{Conclusion}
In this paper, we unify recent efforts on combining cutting-edge semi-supervised learning, contrastive learning, and noisy label learning together. We propose CSSL and CoDiM, which leverage contrastive learning not only to provide self-supervised pre-training but also to further provide consistency regularization besides classical semi-supervised learning processes. We evaluate our methods through extensive experiments on multiple benchmarks across many datasets and show that CoDiM steadily outperforms state-of-the-art methods. Through this work, we address the new possibilities to combine popular methods in different weakly supervised learning fields together and will then explore new ways to accelerate and strengthen the fusion of these methods as our future targets. 

\clearpage
\bibliography{aaai22}

\clearpage
\appendix
\section{A: Implementation details}
Both CSSL and CoDiM contain two phases: a SelfCon pre-training and a contrastive semi-supervised learning style process, except that CoDiM maintains two networks and an iterative Gaussian Mixture Model (GMM)-based clean/noisy data separation within the DivideMix framework. Note that without loss of generality, we keep the same for the shared parameters of CSSL and CoDiM. Also for all experiments, training samples are sampled randomly without replacement.
\subsection{CIFAR-10/100}
In the first phase, we use an 18-layer PreAct Resnet as the network backbone with a 2-layer projection head. The dimensions of hidden and output layers of the projection head are both 256. The input size is 32 $\times$ 32. SimCLR is used to conducted self pre-training. The model is optimized using SGD with a batch size of 512. The weight decay and momentum are set as 0.0005 and 0.9, respectively. We train the model for 800 epochs. In the first 10 epochs, the learning rate gradually increases from 0 to 0.06 and then decreases to 0 at the last epoch in a Cosine Annealing manner. The temperature $\tau_1$ when computing the contrastive loss is set to 0.5. Following DivideMix \cite{li2020dividemix} and AugDesc-WS \cite{nishi2021augmentation}, we keep most parameters in the second phase unchanged. The backbone and the projection layers are initialized with pre-trained parameters in the first layer, and a 2-layer classification head is randomly initialized. The whole model is firstly warmed up for 10 epochs for CIFAR-10 (except that we set it to 1 for `CIFAR-10 with 90\% symmetric noise') and 30 epochs for CIFAR-100, with a batch size of 128 and then optimized with SelfCon/SupCon loss and SSL loss with a batch size of 512. The temperature $\tau_2$ of SelfCon is 0.5 and the temperature $\tau_{3}$ of SupCon is 0.07. The weak augmentation involves random crop and horizontal flip. The strong augmentation used is AutoAugment following AugDesc-WS. The initial learning rate is 0.02 for all settings except that for settings of `CIFAR-10 with 90\% symmetric noise' and `CIFAR-100 with 90\% symmetric noise', the initial learning rate is 0.002. The total training epochs are 300 for `CIFAR-10 with 20\% and 50\% symmetric noise', 350 epochs for `CIFAR-10 with 80\% and 90\% symmetric noise, and 40\% asymmetric noise', and 400 epochs for all `CIFAR-100' experiments. The learning rate drops to 10\% of the original value when running for roughly half of the total epochs. The SelfCon/SupCon loss weight is set to 1 for all experiments except 0.1 for `CIFAR-10 with 80\% symmetric noise' and 0.01 for `CIFAR-10 with 40\% asymmetric noise'.

\subsection{WebVision}
In the first phase, we use an inception-resnet v2 as the network backbone with a 2-layer projection head. The  dimensions of hidden and output layers of the projection head are both 256. The input size is 299 $\times$ 299. SimCLR is used to conducted self pre-training. The model is optimized using SGD with a batch size of 256. The weight decay and momentum are set as 0.0005 and 0.9, respectively. The model is trained for 300 epochs. The learning rate increases from 0 to 0.01 in the first 10 epochs and decreases to 0 eventually via Cosine Annealing. The temperature $\tau_1$ when computing the contrastive loss is set to 0.5. In the second phase, the model is initialized similarly to that of CIFAR-10/100 experiments. The model is warmed up for 1 epoch with a batch size of 64, then optimized in a multitask way with a batch size of 32. The temperature $\tau_2$ of SelfCon is 0.5 and the temperature $\tau_{3}$ of SupCon is 0.07. The total number of epochs is 100. The initial learning rate is 0.01 and decreased to 0.001 at the 50-$th$ epoch. We apply SupCon loss on the clean subset with the weight of 0.1. The weak augmentation includes resize, random crop and horizontal flip. The strong augmentation used is AutoAugment following AugDesc-WS. 

\subsection{Animal-10N}
We apply VGG19 with batch normalization as the network backbone with a 2-layer projection head. The dimensions of hidden and output layers of the projection head are both 256. The input size is 64 $\times$ 64. SimCLR is used to conducted self pre-training. The model is optimized using SGD with a batch size of 1024. The weight decay and momentum are set as 0.0005 and 0.9. The model is trained for 300 epochs. The learning rate increases from 0 to 0.12 in the first 10 epochs and decreases to 0 in the end by Cosine Annealing. The temperature $\tau_1$ when computing the contrastive loss is set to 0.5. In the second phase, the model shares the pre-trained parameters as initialization. The model is warmed up for 5 epochs with a batch size of 256, then optimized in a multitask manner with a batch size of 128. The temperature $\tau_2$ of SelfCon is 0.5 and the temperature $\tau_{3}$ of SupCon is 0.07. The model is trained for 100 epochs with an initial learning rate of 0.01. The learning rate is divided by 5 at the 50-$th$ and 75-$th$ epochs, respectively. SupCon loss is computed on the clean subset and the weight is set to 1. The weak augmentation contains random crop and horizontal flip. The strong augmentation used is AutoAugment following AugDesc-WS. 

\section{B: Full ablation studies on CSSL}
\begin{table}[h]
\begin{center}
\begin{tabular}{lcccccccccc}
\hline
Dataset & & \multicolumn{2}{c}{CIFAR-10} \\
\hline
Methods/labeled ratio & & 20\% & 80\%  \\
\hline
\textbf{w/o SelfCon pre-training} \\
\hline
\multirow{2}{*}{SSL} & Best & 89.3 & 96.2 \\
& Last & 89.0 & 96.0 \\
\hline
\multirow{2}{*}{SSL-L(Self)-U(Self)} & Best & 89.1 & 96.2 \\
& Last & 88.9 & 96.0 \\
\hline
\multirow{2}{*}{SSL-L(Sup)-U(Self)} & Best & 91.7 & 96.6 \\
& Last & 91.5 & 96.4 \\
\hline
\hline
\multirow{2}{*}{SSL-L(Sup) Only} & Best & 90.4 & 96.5 \\
& Last & 90.1& 96.3 \\
\hline
\multirow{2}{*}{SSL-L(Self) Only} & Best & 89.0 & 96.2 \\
& Last & 88.6& 96.0 \\
\hline
\multirow{2}{*}{SSL-U(Self) Only} & Best & 89.1& 95.9 \\
& Last & 89.0 & 95.6\\
\hline
\textbf{w/ SelfCon pre-training} \\
\hline
\multirow{2}{*}{SSL} & Best & 94.0 & 96.8 \\
& Last & 93.9 & 96.7 \\
\hline
\multirow{2}{*}{SSL-L(Self)-U(Self)} & Best & \underline{94.6} & 96.6 \\
& Last & \underline{94.4} & 96.5 \\
\hline
\multirow{2}{*}{SSL-L(Sup)-U(Self)\textbf{(CSSL)}} & Best & \textbf{94.7} & \underline{96.9} \\
& Last & \textbf{94.4} & \underline{96.8} \\
\hline
\hline
\multirow{2}{*}{SSL-L(Sup) Only} & Best & 94.1 & \textbf{97.1} \\
& Last & 93.8 & \textbf{96.9} \\
\hline
\multirow{2}{*}{SSL-L(Self) Only} & Best & 94.4 & 96.6 \\
& Last & 94.3 & 96.5 \\
\hline
\multirow{2}{*}{SSL-U(Self) Only} & Best & 94.5 & 96.5 \\
& Last & 94.3 & 96.3 \\
\hline
\end{tabular}
\end{center}
\caption{Comparing CSSL with basic SSL algorithm in terms of test accuracy (full table). \textit{S}(\textit{alg}) denotes applying \textit{alg} on samples drew from set \textit{S} in the second phase. \textit{S} $\in$\{\textbf{L}abeled,\textbf{U}nlabeled\}, \textit{alg} $\in$ \{\textbf{Self}Con, \textbf{Sup}Con\}. We find that CSSL achieves competitive performance on different levels of labeled ratio. We also find that when given many labels, only applying SupCon to labeled set is useful, while given less labels, it's good to apply SelfCon to unlabeled set. Here bold number means the best one, and underlined number means the best runner-up one in each setting. }
\label{tab:table1_complete}
\end{table}

\begin{table}[ht]
\begin{center}
\begin{tabular}{lccccc}
\hline
\multirow{2}{*}{Dataset} & Back- & & \multicolumn{2}{c}{\multirow{2}{*}{CIFAR-10}} \\
& Bone & & &\\
\hline
Noise type & & & Sym. & Asym. \\
\hline
Methods/Noise ratio & & & 50\% & 40\%  \\
\hline
\multirow{2}{*}{CoDiM-Sup \textbf{(Ours)}} & PreAct & Best & \textbf{96.5} & 95.2 \\
& Res18 & Last & \textbf{96.3} & 95.0 \\
\hline
\multirow{2}{*}{CoDiM-Self \textbf{(Ours)}} & PreAct & Best & 95.3 & \textbf{95.5} \\
& Res18 & Last & 95.2 & \textbf{95.3} \\
\hline
C2D (DivideMix w/& PreAct & Best & 95.5 & 93.6 \\
SelfCon pre-training)& Res18 & Last & 95.3 & 93.3 \\
\hline
CoDiM-Sup & PreAct & Best & 95.5 & 94.1 \\
(w/o pre-training) & Res18 & Last & 95.4 & 93.9 \\
\hline
CoDiM-Self & PreAct & Best & 94.8 & 94.4 \\
(w/o pre-training) & Res18 & Last & 94.6 & 94.3 \\
\hline
\multirow{2}{*}{DivideMix} & PreAct & Best & 95.4 & 94.6 \\
& Res18 & Last & 95.1 & 94.3 \\
\hline
\hline
\multirow{2}{*}{CoDiM-Sup} &Basic  & Best & \underline{89.3} & 88.2 \\
& Res18 & Last & \underline{89.0} & 88.0 \\
\hline
\multirow{2}{*}{CoDiM-Self} &Basic  & Best & 88.8 & \underline{88.5} \\
& Res18 & Last & 88.6 & \underline{88.3} \\
\hline
C2D (DivideMix w/ & Basic & Best & 88.0 & 88.0 \\
SelfCon pre-training)& Res18 & Last & 87.9 & 87.8 \\
\hline
CoDiM-Sup (Image- & Basic & Best & 89.2 & 88.0 \\
Net pre-training)&  Res18 & Last & 89.0 & 87.8 \\
\hline
CoDiM-Self (Image- & Basic & Best & 88.5 & 87.7 \\
Net pre-training)&  Res18 & Last & 88.4 & 87.5 \\
\hline
DivideMix (Image- & Basic & Best & 87.6 & 88.0 \\
Net pretraining) &  Res18 & Last & 87.4 & 87.9 \\
\hline
CoDiM-Sup & Basic  & Best & 87.5 & 86.7 \\
(w/o pre-training) & Res18 & Last & 87.3 & 86.4 \\
\hline
CoDiM-Self & Basic  & Best & 87.2 & 87.0 \\
(w/o pre-training) & Res18 & Last & 87.0 & 86.9 \\
\hline
\multirow{2}{*}{DivideMix} & Basic & Best & 84.4 & 86.6 \\
& Res18 & Last & 84.2 & 86.4 \\
\hline
\end{tabular}
\end{center}
\caption{Ablation study results on effects of network architecture and pre-training. Note, C2D can be seen as using SelfCon pre-training and then apply DivideMix. Here we didn't provide results for using ImageNet pre-training on PreAct Resnet-18 as it's not publicly accessible. We find that our results are independent with the option of different network architectures of feature extractor. We also find that using SelfCon pre-training is always a better choice. Here bold and underlined numbers mean the best ones using PreAct Resnet-18 and basic Resnet-18.}
\label{tab:table6_app}
\end{table}

In Table~\ref{tab:table1_complete}, we show the full results of experiments conducted in main paper (Table 1 in main paper). Besides the results shown in main paper, we test three more alternatives (only apply SupCon/SelfCon on labeled set, or only apply SelfCon on unlabeled set) to combine contrastive learning with semi-supervised learning. Our first key observation is that all trials benefit from leveraging SelfCon pre-training. Note that the improvements are much more obvious when given less labeled samples. Our second observation is that, CSSL achieves the best given 20\% labeled samples and best runner-up given 80\% labeled samples, which provides competitive performances on different levels of labeled ratio. Our third observation is that when the labeled ratio is high, it's actually useful to only apply SupCon on the labeled set, as it achieves the best given 80\% of the labels. This also empirically supports our findings that CoDiM-Sup can acquire improvements under a low ratio of label noise. Also, we can see that, when given fewer labels, it's beneficial to apply SelfCon on the unlabeled set. However, when dealing with label noise is that, as the result of GMM can not fully specify label noise, it's actually harmful to further apply SelfCon on a possibly noisy set (As shown in Table 2, main paper).
\section{C: Ablation studies on CIFAR-10/100.}
\subsection{Choices of pre-training and network architecture}
In this section, we evaluate the effects of using different pre-trainings and network architectures. We conduct experiments under two settings: CIFAR-10 with 50\% symmetric noise and 40\% asymmetric noise. We evaluate two network backbones: PreAct Resnet-18 and Basic Resnet-18, which are both common choices when mining from CIFAR-10. We also evaluate 3 pre-training options: using SelfCon pre-training, ImageNet pre-training, and no pre-training. Since public ImageNet pre-training for PreAct Resnet-18 is not accessible, we only evaluate this option on the basic Resnet-18 backbone. As C2D is simply to leverage SelfCon pre-training before applying DivideMix, we use it to notify this setting. Table~\ref{tab:table6_app} shows the results of these experiments. Note here we are not interested in the average gap between using PreAct Resnet-18 and Basic Resnet-18, and only want to see the differences between experiment pairs when only one option is adjusted.
Firstly, we find that our methods achieve relatively best and consistent performance when using the same backbone, as CoDiM-Sup wins in 50\% symmetric noise case and CoDiM-Self wins in 40\% asymmetric noise case. Secondly, we find that using SelfCon pre-training is always a better choice, regardless of which backbone is used.

\subsection{Label correction of 90\% noise ratio on CIFAR-100}
In this section, we notify a customized label correction step used when dealing with 90\% symmetric noise on CIFAR-100. Note this technique is applied to C2D as well, in order to make fair comparisons. Also, as REED already contains a more complex label correction stage, we follow the original setting when re-implementing REED. The idea of this label correction step is simple. During the `warming-up' stage (just after pre-training using SelfCon), we copy and fix the pre-trained weights to the feature extractor, and train the classifier head with traditional CE-loss using all data with noisy labels for 100 epoch, using SGD optimizer with a learning rate of 0.005. The weight decay and momentum are set as 0.0005 and 0.9, respectively. Then, we directly utilize the predictions of the classifier and change all the labels to the class which the classifier outputs with the largest probabilities. Then, we randomly re-set the weight of the classifier head and start the second phase of CoDiM or other algorithms like DivideMix. Note here we do not re-set the weights used in feature extractor, but make it changeable again. We evaluate the effect of this small process, and show the result in Table~\ref{tab:table7_app}. We note that this label correction step is crucial to our methods.

\subsection{Ablation studies on augmentation strategies and other technologies}
In this section, we provide more results of ablation studies on CIFAR-10 and CIFAR-100, as shown in Table \ref{tab:tableC1}. We see that though altering augmentation strategies of CoDiM-Sup can provide even better results, CoDiM-Sup provides more robust and competitive results across all cases. Note here we see that under relatively low ratio of symmetric noise, using different augmentation strategies might be a good choice, this indicates more efforts are needed on specifying better augmentation strategies. 
\begin{table*}[ht]
\begin{center}
\begin{tabular}{lcccccccccc}
\hline
Dataset & & \multicolumn{5}{c}{CIFAR-10} & \multicolumn{4}{c}{CIFAR-100}\\
\hline
Noise type & & \multicolumn{4}{c}{Sym.} & Asym. & \multicolumn{4}{c}{Sym.} \\
\hline
Methods/Noise ratio & & 20\% & 50\% & 80\% & 90\% & 40\% & 20\% & 50\% & 80\% & 90\% \\
\hline
\multirow{2}{*}{CoDiM-Sup} & Best & \underline{97.0} & \textbf{96.5} & \textbf{94.7} & \textbf{93.7} & \underline{95.2} & 81.1 & \textbf{78.1} & \textbf{67.3} & \underline{55.2}\\
& Last & \underline{96.9} & \textbf{96.4} & \textbf{94.6} & \textbf{93.4} & \underline{95.0} & 80.9 & \textbf{77.9} & \textbf{67.0} & \underline{54.9} \\
\hline
\multirow{2}{*}{CoDiM-Self} & Best & 96.5 & 95.3 & 94.3 & 92.9 & \textbf{95.5} & 80.4 & 77.2 & \underline{63.6} & \textbf{56.4} \\
& Last & 96.3 & 95.2 & 94.2 & 92.7 & \textbf{95.3} & 80.2 & 77.1 & \underline{63.4} & \textbf{56.1} \\
\hline
CoDiM-CSSL & Best & 96.5 & 96.1 & \underline{94.6} & \underline{93.5} & 94.2 & 80.7 & \underline{78.0} & \underline{63.6} & 54.5 \\
(Add SelfCon on $\mathcal{I}$)& Last & 96.3 & 96.0 & \underline{94.4} & \underline{93.4} & 94.0 & 80.5 & \underline{77.8} & \underline{63.4} & 54.3 \\
\hline
CoDiM-bare & Best & 96.4 & 95.5 & \textbf{94.7} & 93.4 & 94.4 & 80.6 & 77.5 & 60.5 & 52.6 \\
(only pre-training) & Last & 96.3 & 95.3 & \textbf{94.6} & 93.3 & 94.1 & 80.4 & 77.2 & 60.2 & 52.3 \\
\hline
{CoDiM-Sup } & Best & 96.5 & 96.0 & 93.9 & 93.4 & 91.6 & 79.0 & 76.0 & 48.1 & 37.7 \\
(w/o Co-training) & Last & 96.3 & 95.8 & 93.7 & 93.3 & 90.7 & 78.7 & 75.8 & 47.8 & 37.3 \\
\hline
CoDiM-Sup & Best & 96.6 & 95.5 & 92.7 & 48.1 & 94.1 & 80.5 & 77.3 & 59.2 & 40.1 \\
(w/o pre-training)& Last & 96.5 & 95.4 & 92.5 & 47.8 & 93.9 & 80.3 & 77.1 & 59.0 & 40.1 \\
\hline

CoDiM-Sup & Best & \textbf{97.1} & \underline{96.4} & 94.4 & 91.8 & 94.2 & 81.3 & 77.9 & 60.0 & 54.7 \\
(All step use $A_{sa}(.)$)& Last & \textbf{97.0} & \underline{96.3} & 94.3 & 91.6 & 93.9 & 81.1 & 77.7 & 59.8 & 54.6 \\
\hline
CoDiM-Sup & Best & 96.8 & 95.8 & 94.3 & 92.7 & 94.3 & \textbf{81.7} & \textbf{78.1} & 63.5 & 53.7 \\
(SupCon use $A_{wa}(.)$) & Last & 96.7 & 95.6 & 94.1 & 92.5 & 93.9 & \textbf{81.5} & \textbf{77.9} & 63.2 & 53.5 \\
\hline
CoDiM-Sup (Mix-& Best & \textbf{97.1} & 96.4 & 94.0 & 93.3 & 92.8 & \underline{81.5} & 77.6 & 61.3 & 53.7 \\
Match only use $A_{wa}(.)$ )& Last & \textbf{97.0} & 96.2 & 93.9 & 93.2 & 90.6 & \underline{81.4} & 77.2 & 61.0 & 53.5 \\
\hline
\end{tabular}
\end{center}
\caption{Ablation study results in terms of test accuracy ($\%$) on CIFAR-10 and CIFAR-100.}
\label{tab:tableC1}
\end{table*}

\begin{table}[ht]
\begin{center}
\begin{tabular}{lcccccccccc}
\hline
Dataset & & \multicolumn{2}{c}{CIFAR-100} \\
Noise Type \& Ratio & & \multicolumn{2}{c}{Sym. / 90\%}\\
\hline
Methods/Label correction& & No &Yes \\
\hline
\multirow{2}{*}{C2D} & Best & 39.4 & 40.1 \\
& Last & 39.1 & 40.1 \\
\hline
\multirow{2}{*}{CoDiM-bare} & Best & 39.4 & 40.1 \\
& Last & 39.1 & 40.1 \\
\hline
\multirow{2}{*}{CoDiM-Sup} & Best & 45.2 & 55.2 \\
& Last & 45.0 & 54.9 \\
\hline
\multirow{2}{*}{CoDiM-Self} & Best & 48.6 & 56.4 \\
& Last & 48.4 & 56.1 \\
\hline
\end{tabular}
\end{center}
\caption{The effect of label correction step used in 90\% symmetric noise settings on CIFAR-100. }
\label{tab:table7_app}
\end{table}

\section{D: Full Comparison with existing methods on CIFAR-10/100.}
In this section, we provide the full version table (Table \ref{tab:table2_complete}) showing the results on CIFAR-10 and CIFAR-100 (as noticed in Table 2, main paper). Note that we already provide the most recent and important baselines in the main paper. We also provide more results on 40\% asymmetric noise on CIFAR-10 setting in Table \ref{tab:table2_complete2_asym}.

\begin{table}[h]
\begin{center}
\begin{tabular}{lccc}
\hline
Method & Best & Last \\
\hline
Cross-Entropy &  85.0 & 72.3 \\
F-correction \cite{patrini2017making} & 87.2 & 83.1 \\
M-correction \cite{arazo2019unsupervised} & 87.4 & 86.3 \\
Iterative-CV \cite{chen2019understanding} & 88.6 & 88.0 \\
P-correction \cite{yi2019probabilistic} & 88.5 & 88.1 \\
Joint-Optim \cite{tanaka2018joint} & 88.9 & 88.4 \\
Meta-Learning \cite{li2019learning} & 89.2 & 88.6 \\
PENCIL & 91.2& - \\
Distilling & 90.2 & - \\
REED \cite{zhang2020decoupling} & 92.4 & 92.3 \\
DivideMix \cite{li2020dividemix} & 93.4 & 92.1 \\
C2D \cite{zhang2020decoupling} & 93.6 & 93.3 \\
DM-AugDesc \cite{nishi2021augmentation} & 94.6 & 94.3 \\
\hline
CoDiM-CSSL & 94.2 & 94.0 \\
CoDiM-bare& 94.4 & 94.1 \\

\textbf{CoDiM-Sup (Ours)} & \underline{95.2} & \underline{95.0} \\
\textbf{CoDiM-Self (Ours)} & \textbf{95.5} & \textbf{95.3} \\
\hline
\end{tabular}
\end{center}
\caption{Comparison with existing methods on CIFAR-10 with 40\% asymmetric noise.}
\label{tab:table2_complete2_asym}
\end{table}

\begin{table*}[h]
\begin{center}
\begin{tabular}{lcccccccccc}
\hline
\multicolumn{2}{c}{Dataset} & \multicolumn{4}{c}{CIFAR-10} & \multicolumn{4}{c}{CIFAR-100} \\
\hline
Noise type & & \multicolumn{4}{c}{Sym.}& Asym.  & \multicolumn{4}{c}{Sym.} \\
\hline
\multicolumn{2}{c}{Method/Noise ratio} & 20\% & 50\% & 80\% & 90\% & 40\% & 20\% & 50\% & 80\% & 90\% \\
\hline 
\multirow{2}{*}{Cross-Entropy} & Best & 86.8 & 79.4 & 62.9 & 42.7 & 85.0 & 62.0 & 46.7 & 19.9 & 10.1 \\
 & Last & 82.7 & 57.9 & 26.1 & 16.8 & 72.3 & 61.8 & 37.3 & 8.8 & 3.5 \\
\multirow{2}{*}{Bootstrap \cite{reed2014training}} & Best & 86.8 & 79.8 & 63.3 & 42.9 & - & 62.1 & 46.6 & 19.9 & 10.2 \\
 & Last & 82.9 & 58.4 & 26.8 & 17.0 & - & 62.0 & 37.9 & 8.9 & 3.8 \\
\multirow{2}{*}{F-correction \cite{patrini2017making}} & Best & 86.8 & 79.8 & 63.3 & 42.9 & 87.2 & 61.5 & 46.6 & 19.9 & 10.2 \\
 & Last & 83.1 & 59.4 & 26.2 & 18.8 & 83.1 & 61.4 & 37.3 & 9.0 & 3.4 \\
\multirow{2}{*}{Co-teaching+ \cite{yu2019does}} & Best & 89.5 & 85.7 & 67.4 & 47.9 & - & 65.6 & 51.8 & 27.9 & 13.7 \\
 & Last & 88.2 & 84.1 & 45.5 & 30.1 & - & 64.1 & 45.3 & 15.5 & 8.8 \\
\multirow{2}{*}{Mixup \cite{zhang2017mixup}} & Best & 95.6 & 87.1 & 71.6 & 52.2 & - & 67.8 & 57.3 & 30.8 & 14.6 \\
 & Last & 92.3 & 77.6 & 46.7 & 43.9 & - & 66.0 & 46.6 & 17.6 & 8.1 \\
\multirow{2}{*}{P-correction \cite{yi2019probabilistic}} & Best & 92.4 & 89.1 & 77.5 & 58.9 & 88.5 & 69.4 & 57.5 & 31.1 & 15.3 \\
 & Last & 92.0 & 88.7 & 76.5 & 58.2 & 88.1 & 68.1 & 56.4 & 20.7 & 8.8 \\
\multirow{2}{*}{Meta-Learning \cite{li2019learning}} & Best & 92.9 & 89.3 & 77.4 & 58.7 & 89.2 & 68.5 & 59.2 & 42.4 & 19.5 \\
 & Last & 92.0 & 88.8 & 76.1 & 58.3 & 88.6 & 67.7 & 58.0 & 40.1 & 14.3 \\
\multirow{2}{*}{M-correction \cite{arazo2019unsupervised}} & Best & 94.0 & 92.0 & 86.8 & 69.1 & 87.4 & 73.9 & 66.1 & 48.2 & 24.3 \\
 & Last & 93.8 & 91.9 & 86.6 & 68.7 & 86.3 & 73.4 & 65.4 & 47.6 & 20.5 \\
\multirow{2}{*}{DivideMix \cite{li2020dividemix}} & Best & 96.1 & 94.6 & 93.2 & 76.0 & 93.4 & 77.3 & 74.6 & 60.2 & 31.5 \\
 & Last & 95.7 & 94.4 & 92.9 & 75.4 & 92.1 & 76.9 & 74.2 & 59.6 & 31.0 \\
\multirow{2}{*}{DM-AugDesc \cite{nishi2021augmentation}} & Best & 96.3 & 95.4 & 93.8 & 91.9 & 94.6 & 79.5 & 77.2 & 66.4 & 41.2 \\
 & Last & 96.2 & 95.1 & 93.6 & 91.8 & 94.3 & 79.2 & 77.0 & 66.1 & 40.9 \\
\multirow{2}{*}{C2D\cite{zheltonozhskii2021contrast} } & Best & 96.4 & 95.5 & 94.3 & 93.2 & 93.6 & 78.7 & 76.8 & 59.8 & 52.1 \\
 & Last & 96.3 & 95.3 & 94.2 & 93.0 & 93.3 & 78.4 & 76.4 & 59.6 & 51.9 \\
\multirow{2}{*}{REED \cite{zhang2020decoupling}} & Best & 95.9 & 95.4 & 94.4 & \underline{93.6} & 92.4 & 76.9 & 72.8 & 65.6 & \underline{55.7} \\
 & Last & 95.8 & 95.2 & 94.2 & \underline{93.5} & 92.3 & 76.7 & 72.5 & 65.4 &  \underline{55.2} \\
\hline
\hline
CoDiM-bare & Best & 96.4 & 95.5 & \underline{94.6} & 93.4 & 94.4 & 80.6 & 77.5 & 60.5 & 52.6 \\
(only pre-training) & Last & 96.3 & 95.3 & \underline{94.5} & 93.3 & 94.1 & 80.4 & 77.2 & 60.2 & 52.3 \\
\hline
CoDiM-CSSL & Best & \underline{96.5} & \underline{96.1} & 94.6 & 93.5 & 94.2 & \underline{80.7} & \underline{78.0} & \underline{63.6} & 54.5 \\
(add SelfCon on $\mathcal{I}$)& Last & \underline{96.3} & \underline{96.0} & 94.4 & 93.4 & 94.0 & \underline{80.5} & \underline{77.8} & \underline{63.4} & 54.3 \\
\hline
\multirow{2}{*}{\textbf{CoDiM-Self (Ours)}} & Best & \underline{96.5} & 95.3 & 94.3 & 92.9 & \textbf{95.5} & 80.4 & 77.2 & \underline{63.6} & \textbf{56.4} \\
& Last & \underline{96.3} & 95.2 & 94.2 & 92.7 & \textbf{95.3} & 80.2 & 77.1 & \underline{63.4} & \textbf{56.1} \\
\hline
\multirow{2}{*}{\textbf{CoDiM-Sup (Ours)}} & Best & \textbf{97.0} & \textbf{96.5} & \textbf{94.7} & \textbf{93.7} & \underline{95.2} & \textbf{81.1} & \textbf{78.1} & \textbf{67.3} & 55.2 \\
 & Last & \textbf{96.9} & \textbf{96.4} & \textbf{94.6} & \textbf{93.4} & \underline{95.0} & \textbf{80.9} & \textbf{77.9} & \textbf{67.0} & 54.9 \\
\hline
\end{tabular}
\end{center}
\caption{Comparison with existing methods on CIFAR-10/100 with different noise settings (full table). We re-implement C2D and REED here. Note CoDiM-bare can be regarded as a combination of SelfCon pre-training and then apply DM-AugDesc. CoDiM-CSSL apply SupCon on $\mathcal{C}$ and SelfCon on $\mathcal{I}$ following CSSL. Note, we leverage the label correction step discussed in the earlier section for all experiments on 90\% symmetric noise on CIFAR-100 we re-implemented except for REED.}
\label{tab:table2_complete}
\end{table*}

\section{E: Full Comparison with existing methods on Webvision.}
Here we provide full version table of results on WebVision (Table \ref{tab:table4_complete}). Note that we already provide important baselines in the table in main paper.
\begin{table*}[!htbp]
\begin{center}
\begin{tabular}{lcccc}
\hline
\multirow{2}{*}{Method} & \multicolumn{2}{c}{WebVision} & \multicolumn{2}{c}{ILSVRC12} \\
 & top1 & top5 & top1 & top5 \\
\hline
F-correction \cite{patrini2017making} & 61.12 & 82.68 & 57.36 & 82.36 \\
Decoupling \cite{malach2017decoupling} & 62.54 & 84.74 & 58.26 & 82.26 \\
D2L \cite{ma2018dimensionality} & 62.68 & 84.00 & 57.80 & 81.36 \\
MentorNet \cite{jiang2018mentornet} & 63.00 & 81.40 & 57.80 & 79.92 \\
Co-teaching \cite{han2018co} & 63.58 & 85.20 & 61.48 & 84.70 \\
Iterative-CV \cite{chen2019understanding} & 65.24 & 85.34 & 61.60 & 84.98 \\
DivideMix \cite{li2020dividemix} & 77.32 & 91.64 & 75.20 & 90.84 \\
ELR+ \cite{liu2020early} & 77.78 & 91.68 & 70.29 & 89.76 \\
LongReMix \cite{cordeiro2021longremix} & 78.92 & 92.32 & - & - \\
DM-AugDesc\cite{nishi2021augmentation} & 78.64 & 93.20 & 75.52 & 92.12 \\ 
GJS \cite{englesson2021generalized} & 79.28 & 91.22 & 75.50 & 91.27 \\
C2D\cite{zheltonozhskii2021contrast} & 80.20 & 93.22 & 76.64 & 92.32 \\ 
\hline
CoDiM-bare  & 80.32 & \underline{93.40} & 76.60 & 92.36 \\
CoDiM-Self & 80.12 & 93.52 & \textbf{77.24} & \underline{92.48}\\ 
CoDiM-Sup & \textbf{80.88} & 92.48 & 76.52 & 91.96 \\
\hline
\end{tabular}
\end{center}
\caption{Comparison with existing methods on WebVision. We re-implement DM-AugDesc and C2D.}
\label{tab:table4_complete}
\end{table*}

\clearpage
\section{F: more t-SNE results}
Here we provide t-SNE visualizations of learning with noisy labels for different experimental settings on the test set of CIFAR-10. We first show visualizations of self-supervised pre-training. Then CoDiM-Sup-bare (CoDiM-Sup without self pre-training), CoDiM-Self and CoDiM-Sup are shown on settings of `20\% symmetric noise, 50\% symmetric noise, 80\% symmetric noise, 90\% symmetric noise, and 40\% asymmetric noise', respectively. We can basically observe that experiments on high-ratio noise benefit more from self pre-training. Even with the noise ratio as high as 90\%, the models learned by our method still cluster test samples well. 

\begin{figure*}[h]
\centering
{
\includegraphics[width=0.3\linewidth]{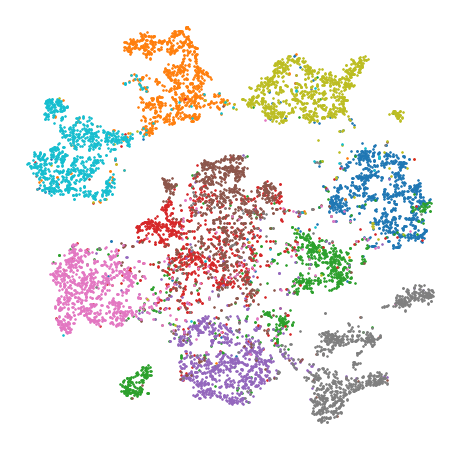}
}
\caption{Self pre-training}
\end{figure*}

\begin{figure*}[h]
\centering
\subfigure[CoDiM-Sup-bare]{
\includegraphics[width=0.3\linewidth]{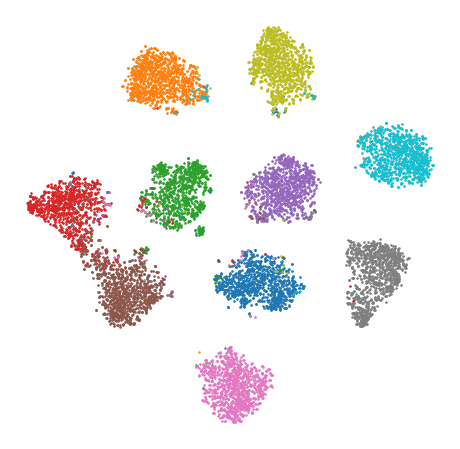}
}
\quad
\subfigure[CoDiM-Self]{
\includegraphics[width=0.3\linewidth]{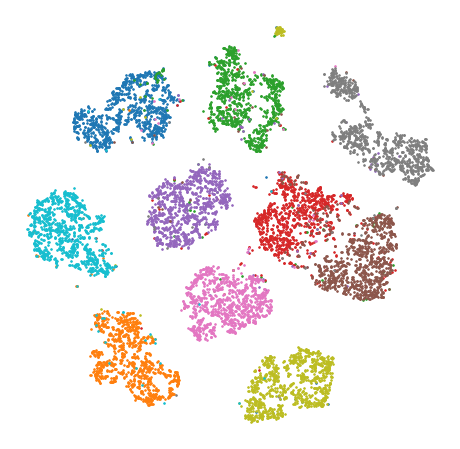}
}
\quad
\subfigure[CoDiM-Sup]{
\includegraphics[width=0.3\linewidth]{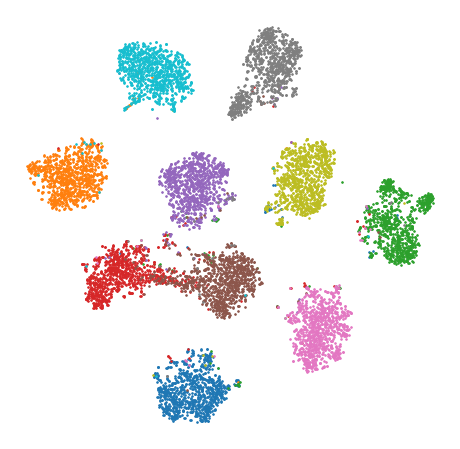}
}
\caption{t-SNE visualizations of experiments on 20\% Symmetric Noise.}
\end{figure*}

\begin{figure*}[h]
\centering
\subfigure[CoDiM-Sup-bare]{
\includegraphics[width=0.3\linewidth]{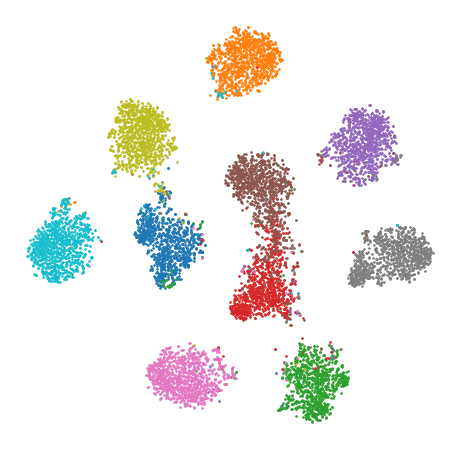}
}
\quad
\subfigure[CoDiM-Self]{
\includegraphics[width=0.3\linewidth]{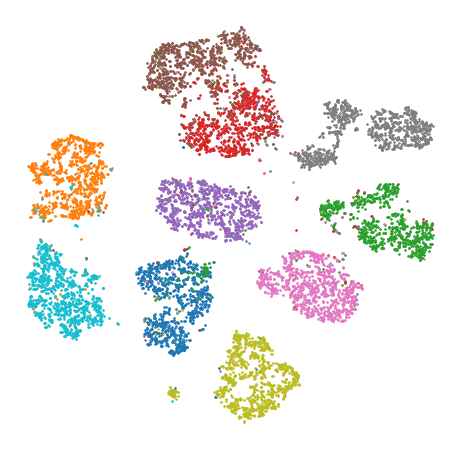}
}
\quad
\subfigure[CoDiM-Sup]{
\includegraphics[width=0.3\linewidth]{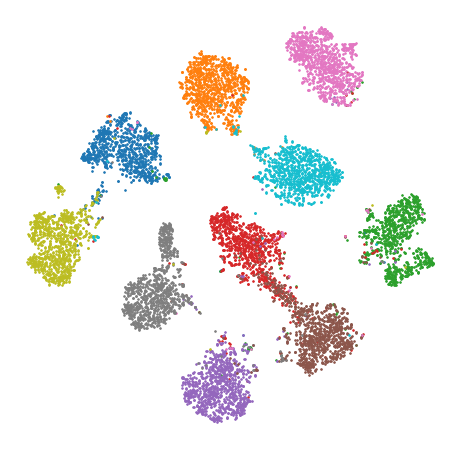}
}
\caption{t-SNE visualizations of experiments on 50\% Symmetric Noise.}
\end{figure*}

\begin{figure*}[h]
\centering
\subfigure[CoDiM-Sup-bare]{
\includegraphics[width=0.3\linewidth]{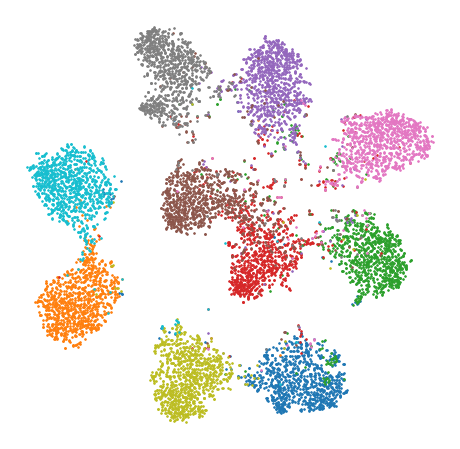}
}
\quad
\subfigure[CoDiM-Self]{
\includegraphics[width=0.3\linewidth]{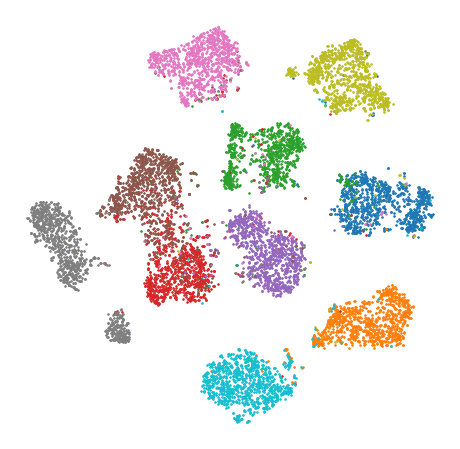}
}
\quad
\subfigure[CoDiM-Sup]{
\includegraphics[width=0.3\linewidth]{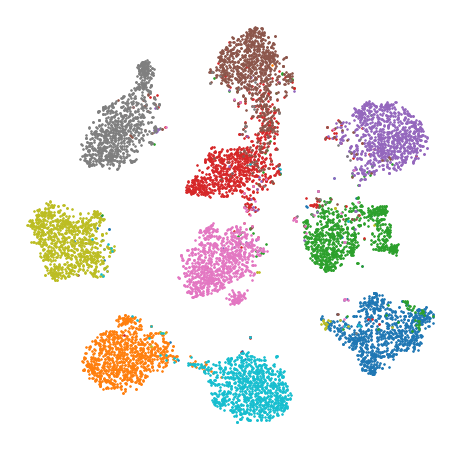}
}
\caption{t-SNE visualizations of experiments on 80\% Symmetric Noise.}
\end{figure*}

\begin{figure*}[h]
\centering
\subfigure[CoDiM-Sup-bare]{
\includegraphics[width=0.3\linewidth]{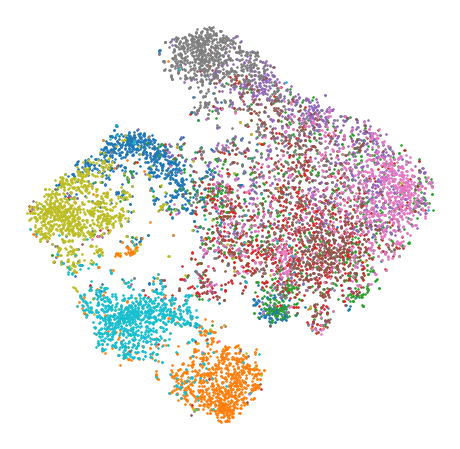}
}
\quad
\subfigure[CoDiM-Self]{
\includegraphics[width=0.3\linewidth]{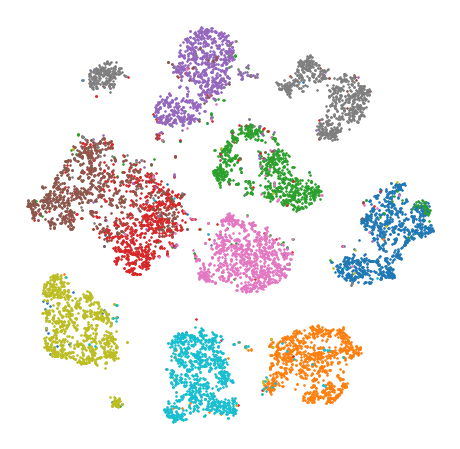}
}
\quad
\subfigure[CoDiM-Sup]{
\includegraphics[width=0.3\linewidth]{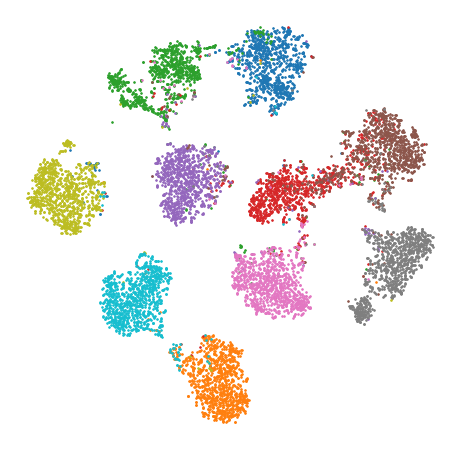}
}
\caption{t-SNE visualizations of experiments on 90\% Symmetric Noise.}
\end{figure*}

\begin{figure*}[h]
\centering
\subfigure[CoDiM-Sup-bare]{
\includegraphics[width=0.3\linewidth]{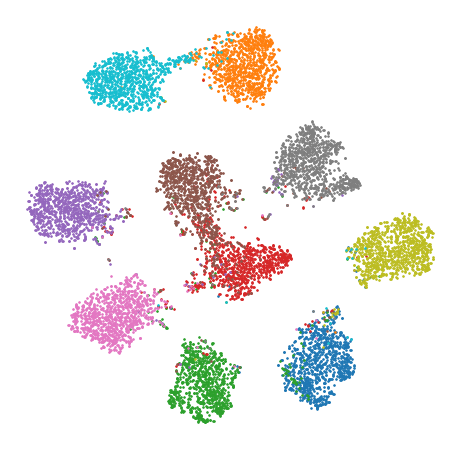}
}
\quad
\subfigure[CoDiM-Self]{
\includegraphics[width=0.3\linewidth]{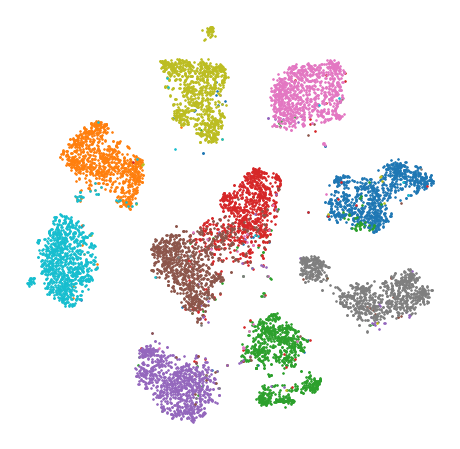}
}
\quad
\subfigure[CoDiM-Sup]{
\includegraphics[width=0.3\linewidth]{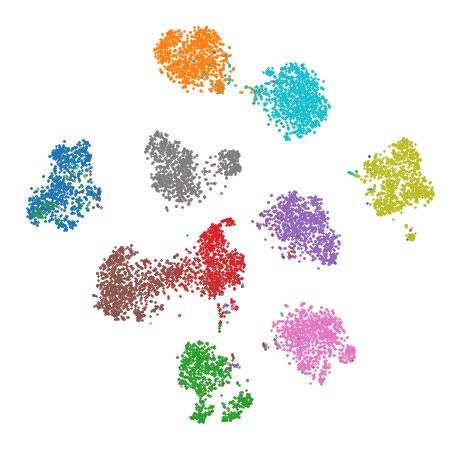}
}
\caption{t-SNE visualizations of experiments on 40\% Asymmetric Noise.}
\end{figure*}

\end{document}